%% file: iclr2024_conference.tex
\documentclass{article} 
\usepackage{iclr2024_conference,times}

\input{math_commands.tex}

\usepackage{hyperref}
\usepackage{url}
\usepackage{algorithm}
\usepackage{algorithmic}
\usepackage{graphicx}
\usepackage{booktabs} 
\usepackage{caption}

\title{ Gradient-Congruity Guided Federated Sparse Training }

\author{Chris Xing Tian, Yibing Liu, Haoliang Li, Ray C.C. Cheung, Shiqi Wang 
\thanks{ Corresponding author} \\
City University of Hong Kong\\
\texttt{xing.tian@my.cityu.edu.hk, lyibing112@gmail.com} \\
\texttt{\{haoliang.li,racheung,shiqwang\}@cityu.edu.hk} \\
}

\newcommand{\bfstart}[1]{\noindent\textbf{#1.}}
\newcommand{\CommetBlue}[1]{{\color[RGB]{45,28,128} #1}}

\iclrfinalcopy 
\begin{document}

\maketitle

\begin{abstract}
Edge computing allows artificial intelligence and machine learning models to be deployed on edge devices, where they can learn from local data and collaborate to form a global model. Federated learning (FL) is a distributed machine learning technique that facilitates this process while preserving data privacy. However, FL also faces challenges such as high computational and communication costs regarding resource-constrained devices, and poor generalization performance due to the heterogeneity of data across edge clients and the presence of out-of-distribution data. In this paper, we propose the Gradient-Congruity Guided Federated Sparse Training (FedSGC), a novel method that integrates dynamic sparse training and gradient congruity inspection into federated learning framework to address these issues. Our method leverages the idea that the neurons, in which the associated gradients with conflicting directions with respect to the global model contain irrelevant or less generalized information for other clients, and could be pruned during the sparse training process. Conversely, the neurons where the associated gradients with consistent directions could be grown in a higher priority. In this way, FedSGC can greatly reduce the local computation and communication overheads while, at the same time, enhancing the generalization abilities of FL. We evaluate our method on challenging non-i.i.d settings and show that it achieves competitive accuracy with state-of-the-art FL methods across various scenarios while minimizing computation and communication costs.
\end{abstract}

\section{Introduction}
\label{sec:intro}
Machine learning has seen significant success in various fields, but traditional centralized training methods pose challenges due to the need for large data sets, high costs, and potential privacy risks. Federated Learning (FL) \cite{yang2019federated,mcmahan2017communication} addresses these issues by allowing multiple parties to collaboratively learn a model without sharing private data. In an FL system, each party trains a local model on their own data, and the weights or gradients are sent to a central server for aggregation. The server updates the global model and sends it back for further training.

FL is ideal for edge computing applications involving distributed devices with limited resources. However, training Deep Neural Networks (DNNs) on these devices presents challenges related to resource efficiency and data heterogeneity. Edge devices may have limited resources for local training and communication with the central server, and they may exhibit heterogeneous data distributions, affecting the generalization performance of FL.

Model compression techniques have been proposed to address resource efficiency, involving the transmission of a compressed parameter/gradient vector between clients and servers \cite{wu2022communication,chen2021communication}. However, this only reduces communication workload and does not create a smaller, more efficient model. The lottery ticket hypothesis \cite{frankle2020linear} suggests that dense neural networks contain sparse subnetworks that can achieve the same accuracy as the original model when trained alone. Some researchers aim to extract a lightweight model from the original model for more efficient client-side training \cite{mugunthan2022fedltn,li2021lotteryfl,tamirisa2023fedselect}, but these methods may impose significant computation and communication overheads and neglect the large data heterogeneity across edge clients.

Regarding the issue of data heterogeneity issue in federated learning, many algorithmic solutions have been proposed in the literature (e.g., \cite{acar2021federated, pmlr-v119-karimireddy20a,li2021model,dai2023tackling}). These strategies add proximal terms to the objective function to regularize local updates with respect to the global model. However, these methods limit local convergence potential and the amount of novel information per communication round, and may not consistently improve performance across different non-IID settings. Some alternative approaches propose personalized federated learning, which trains individual client models, or shares a small subset of an auxiliary dataset with clients to construct a more balanced IID data distribution. However, these methods often overlook client resource constraints and may even incur significant client compute and/or memory overheads.


In this paper, we propose to jointly address the aforementioned limitations with a novel federated learning (FL) framework called Federated Sparse Gradient Congruity (FedSGC). FedSGC integrates dynamic sparse training and gradient congruity inspection \cite{yu2020gradient, mansilla2021domain, tian2023privacy} (i.e., to inspect whether gradient conflicts exist across different data sources)  to enable efficient on-device computation and in-network communication for edge devices in federated learning. To be more specific, we propose a novel prune-and-grow mechanism by examining the gradients across different clients, where client's local training gradients with conflicting directions with respect to the global model contain irrelevant or less generalized information for other clients and could be pruned during the sparse training process, and conversely, gradients with consistent directions with the global model's learning direction could be kept by assigning a higher priority for regrowth of the pruned neurons, thereby fostering the learning of invariant knowledge. 


\section{Related Work}
%

\bfstart{Federated Learning} Federated learning algorithms aim to obtain a global model that minimizes the training loss across all clients. Each client $j$ has a small set of local data $D_j$ for local training, but to preserve user privacy, clients do not share their local data, where the process can be formulated as $\min_{w} f(w) =\sum_{k=1}^{K} p_k F_k(w)$, where $F_k(w)$ denotes the objective of the deep learning model on the $k$-th client, $K$ is the set of clients, $p_k>0$, and $\sum_k p_k=1$. In practice, one can set $p_k = n_k/n$, where $n_k$ and $n$ denote the number of data points in the $k$-th client and the total number of data points, respectively. It is worth noting that federated learning is different from the traditional distributed learning scenario where data partitions are assumed to be i.i.d., meaning that they are generated from the same memoryless stochastic process. However, this assumption does not hold in federated learning. Instead, data can often be heterogeneous among clients.

In the FedAvg \cite{mcmahan2017communication} family of algorithms, training proceeds in communication rounds. At the beginning of each round $r$, the server selects a subset of clients $C^r$,  $C^r \subseteq K$, and sends the current server model parameters $\theta^i$ to the clients. Each client $c \in C^r$ performs $E$ epochs of training using the received model parameters on its local data to produce updated parameters $\theta^{r}_{c}$, which are then uploaded to the server. The server then updates the global model with a weighted average of the sampled clients' parameters to produce $\theta^{r+1}$ for the next round.

\bfstart{Sparse Training} Sparse training performs training and pruning simultaneously by adding a sparsity regularization term to the loss function, producing structured sparsity. Sparse training with dynamic sparsity, also known as Dynamic Sparse Training (DST), is a recent research direction that aims at accelerating the training of neural networks without sacrificing performance. A neural network is initialized with a random sparse topology from scratch. The sparse topology (connectivity) and the weights are jointly optimized during training. During training, the sparse topology is changed periodically through a prune-and-grow cycle, where a fraction of the parameters are pruned, and the same portion is regrown among different neurons. An update schedule determines the frequency of topology updates. Many DST works have been proposed, focusing on improving the performance of sparse training for supervised image classification tasks by introducing different criteria for neuron growth \cite{mostafa2019parameter, evci2020rigging}. DST has demonstrated its success in many fields, such as continual learning \cite{sokar2021spacenet}, feature selection \cite{atashgahi2022quick}, ensembling \cite{liu2021deep}, adversarial training \cite{ozdenizci2021training}, and deep reinforcement learning \cite{sokar2021dynamic}. FedDST \cite{bibikar2022federated}, which is closely related to our research, is based on the foundation of RigL \cite{evci2020rigging}, thereby facilitating computational and communicational efficiency in FL environments. By adopting the gradient magnitude, which was utilized in the original RigL work for directing the pruning and growth process, with mask votes and a sparse weighted average mechanism, it can stabilize learning updates, thereby enhancing the robustness and effectiveness of the global model. However, it does not specifically address the issue of data heterogeneity and exhibits limitations in handling severe data heterogeneity.

\section{Methodology}
We adopt a federated learning setting based on the common FedAvg-like formulation $\min_{w} f(w)$. Instead of sending the full dense parameters $\theta^{r}_{j}$ to the central server at each round $r$, each client $j$ only uploads the sparse parameters along with a bit mask $(\theta^{r}_{j}, m^{r}_{j})$. The bit mask has the same shape as the parameter to indicate whether the parameter at the corresponding index is zero or not.

The central server aggregates the parameters and masks from all clients to produce the global parameters and direction mask $(\theta^{r+1}, d^{r+1})$ for the next round, along with the parameter mask $m^{r+1}$. These are then passed to the selected clients for the next round.
The selected clients use the received parameters and direction mask to perform local sparse training. Following \cite{evci2020rigging}, each client maintains a target overall sparsity $S = \frac{\sum_{l}s^lW^l}{W},  S\in [0,1),$ where $l$ denotes the $l^{th}$ layer of the network and $W^l$ is the number of parameters in that layer. Each layer may have a different layer sparsity $s^l$, which is defined as the ratio of zero parameters to the total number of parameters in that layer, $s^l \in [0,1)$. We adopt Erdös-Rényi Kernel as the layer sparsity distribution ($s^l$) across the network, which is a modified version of Erdös-Rényi formulation \cite{mocanu2018scalable}, to generally allocate higher sparsities to the layers with more parameters while allocating lower sparsities to the smaller ones. 

\subsection{The Prune-and-Grow Mechanism for Sparsity Training}
The dynamic sparse training consists of two phases: learning and re-adjusting. In the learning phase, each client updates the masked parameters using its local training data through standard backpropagation. Parameters with zero mask values remain unchanged and do not join the training.
In the re-adjusting stage, each selected client will periodically re-adjust/update its parameter mask to enable dynamic sparsity. The re-adjusting stage is triggered by the global server: starting from the initial training round, for every $\Delta R$ rounds till $R_{end}$, the global server requests for mask re-adjustment. Each client records its cumulative training epochs since the start of the first round and re-adjusts its mask every $\Delta T$ epochs until it reaches $T^{end}_c$ i.e. the total expected training epochs of the client $c$. The re-adjusting process has two steps: pruning and growing. In the pruning step, for each layer, given the target layer sparsity $s^l$, each client $c$ prunes its layer parameters to a slightly higher sparsity $\overline{s}^l_c=s^l+\sigma_i(1-s^l_c)$, where $\sigma_c$ is a factor that controls the level of over-pruning, i.e., $k=\overline{s}^l_cN^l$ parameters will be pruned (mask set to 0) in total. Inspired by \cite{evci2020rigging}, we also set $\sigma_c$ to be a periodic variable along the federated learning process in our work as 
$\sigma_c = \frac{\alpha}{2}(1+cos(\frac{t^c\pi}{T_c^{end}})), \quad \sigma_c \in [0, \alpha],$
where $\alpha$ is a hyper-parameter and $t_c$ is the current cumulative training epochs of client $c$. This means that each client prunes more parameters at the beginning than in the middle. In the growing step, we need to grow some neurons back (corresponding mask values set from 0 to 1) to achieve the target sparsity level. That means for each layer, we need to grow $\hat{k}$ parameters, where $\hat{k} = (\hat{s}^l - s^l) *N^j$. We will discuss how we select which neurons to grow in the next section. Through the prune-and-grow cycle, the mask space for sparsity neural network training adaptively changes over time.

\subsection{Gradient guided Pruning and Growing}
Our main contribution is to propose a novel criterion for pruning and growing neurons in sparse neural networks. Unlike previous methods (e.g., \cite{atashgahi2022quick}) that grow connections randomly without considering the issue of data heterogeneity, we design our criterion specifically for federated learning settings, where we leverage a global pseudo-gradient\cite{yao2019federated} direction map to optimize the network topology. The global direction map indicates the direction of the change in the global parameters after each round of aggregation, i.e. $d^{r+1}=sign(\theta^{r+1}-\theta^r)$. 
We illustrate the training process of our approach in Algorithm~\ref{alg:fds} in the appendix.

\textbf{Prune criterion. }{On the client-$c$ side, we select the parameters that have local pseudo-gradient of opposite sign from the global direction and then sort them by their weight absolute value in ascending order. We first prune the first $\lambda k$ of them. i.e.}
\begin{equation}
\label{eq:drop_guided}
 K_{guided} = ArgTopK(-|\theta^l_{[i|d^r=-sign({\Delta_c^r})]}|, \lambda k ),
\end{equation}
where we follow \cite{evci2020rigging} to treat the layer parameter as a one-dimensional vector, $i$ is the neuron/parameter index in the vector, and the function $ArgTopK$ returns the indices of the top-k elements in the vector. $\Delta_c^r=\theta^{r,t}_c - \theta^{r,0}_c$ denotes the local pseudo-gradient of client $c$ at the current iteration step $t$ of round $r$, $\lambda$ is a hyper-parameter that controls the proportion of pruning guided by the gradient congruity. The motivation here is that the neurons where the associated global and local gradients are incongruous (i.e., the sign of global and local gradients are different) are less likely to be generalized across heterogeneous data \cite{mansilla2021domain}. For the remaining neurons (i.e., neurons not in $K_{guided}$) to be pruned, we use the magnitude of the parameters only, i.e.,
\begin{equation}
\label{eq:drop_rest}
K_{mag} = ArgTopK(-|\theta^l_{[i \notin K_{guided}]}|, (1-\lambda) k).
\end{equation}
By pruning in this order, we prioritize pruning those with opposite directions while generally pruning by weight magnitude. Finally, we pruned parameters with indices in $K_{guided} \cup K_{mag}$.

\textbf{Grow criterion. } 
{Similar to the pruning stage, we also adopt a prioritized approach to growing neurons. Initially, we grow the neurons that have the largest magnitude of loss gradient, and at the same time, have the same learning direction sign as the global direction, i.e.,}
\begin{equation}
\label{eq:grow_guided}
\hat{K}_{guided}=ArgTopK(|{\nabla}^l_{[i|d^r=sign({\Delta_c^{r,t}})]}|, \lambda \hat{k}),\end{equation}
where we grow the neurons that the associated direction of global and local gradient are congruent (i.e., $i|d^r=sign({\Delta_c^r})$, where $\Delta_c^r=\theta^{r,t}_c - \theta^{r,0}_c$ is the pseudo gradient of client $c$ at current iteration $t$ of round $r$, $i$ is the neuron index). Such neurons are safer to be grown as we expect them to be better generalized across clients with heterogeneous data \cite{mansilla2021domain}.  For the remaining neurons (i.e., neurons not in $K_{guided}$) to be grown, we use the loss gradient magnitude given as
\begin{equation}
\label{eq:grow_rest}\hat{K}_{mag}=ArgTopK(|{\nabla}^l_{[i \notin \hat{K}_{guided}]}|, (1-\lambda) \hat{k}).
\end{equation}
Finally, in the growing step, we grow neurons with indices in $\hat{K}_{guided} \cup \hat{K}_{mag}$.
\subsection{Global Aggregation}
 When the central server receives the sparse parameters and masks from the clients, we perform a sparse weighted average to aggregate them as follows: 
\begin{equation}
\label{eq:central_agg}
\begin{aligned}
&\theta^{r+\frac{1}{2}} = \frac{\sum_{c\in C^r}n_c\theta^r_c m^r_c + n_{rest}\theta^rm^r}{\sum_{c\in C^r}n_cm^r_c+n_{rest}m^r},
&n_{rest} = n - \sum_{c\in C^r}n_c, \\
&\theta^{r+1} = prune \big( ArgTopK(-|\theta^{r+\frac{1}{2}}|, \tilde{k}) \big ),
\end{aligned}
\end{equation}

Different from FedDST, in the central aggregation phase, we still consider the parameters and masks that are held by those clients who do not participate, as we find that updating the global model only by partially selected clients' learning results may lead to unstable performance of the global model. 


In this process, neurons that are zero-masked by all clients get pruned globally. In most cases, the global model sparsity after aggregation may be lower than the target sparsity $S$ since we use OR logic to aggregate the client masks. In such case, we prune some additional neurons with the smallest absolute value of the aggregated weights, i.e. $ArgTopK_i(-|\theta^{r+\frac{1}{2}}|, \tilde{k})$, where $\tilde{k}$ is the number of neurons to be pruned. This again helps us achieve the target sparsity $S$ and obtain the final global parameters $\theta^{r+1}$.

\section{Experiments}
\label{sec:experiments}
In this section, we evaluate our proposed FedSGC on two benchmark datasets: MNIST , CIFAR-10. Our evaluation primarily focuses on comparing FedSGC with two SoTA FL methods that utilize pruning techniques: FedDST~\cite{bibikar2022federated} and PruneFL~\cite{jiang2022model}. We also adapt GraSP~\cite{Wang2020Picking} into FL setting as another baseline. We skip LotteryFL~\cite{li2020lotteryfl} as it is designed for Personalized FL setting. We assess the performance of these methods while considering communication costs.
To reflect a challenging data heterogeneity environment, we create a highly non-IID data distribution among clients, following the same non-iid partition strategy as in FedAvg, where most client only has data from two classes. We also adopt the Dirichlet distribution with parameter $\beta$ to distribute the CIFAR-10 to simulate more relaxing but realistic heterogeneity. 


\subsection{Results on MNIST}
\textbf{Implementation. }{We select 100 clients from the MNIST dataset and randomly choose 10 clients to participate in each round of federated training. All methods are training for 400 federated rounds, 5 local epochs each round. We first sort the MNIST samples by label and split them into 200 shards of size 300. Each client is assigned two shards, resulting in pathological non-IID MNIST dataset partitions, where most clients only have data from two classes.  resulting in a pathological non-IID partition where most clients only have data from two classes. For the federated training settings, we adopt the same CNN architecture and local training algorithm as in FedDST. The CNN consists of two 5x5 convolution layers (the first with 10 channels, the second with 20), a fully connected layer with 50 units and ReLU activation, and a final softmax output layer. The local training is performed via vanilla SGD with a learning rate of $lr=0.001$ and batch size of 50.}

\begin{table*}[t]
	\begin{minipage}{.45\linewidth}
		\centering
		\resizebox{1\columnwidth}{!}{
			\begin{tabular}{lcccc} 
        \toprule
         \textbf{MNIST} &  \multicolumn{4}{c}{Best accuracy encountered at}\\ 
         &  \multicolumn{4}{c}{cumulative upload capacity [MiB]}\\
         Method             &100     &200    &400    &800 \\
          \midrule
         FedAvg             &21.3   &45.2           &73.4           &81.3 \\ 
         FedProx            &\textbf{23.5}    &58.3            &74.3            &82.0 \\ 
         PruneFL            &18.1   &40.2           &60.7           &75.3    \\ 
         GraSP              &20.2   &42.4           &66.1           &70.3    \\ 
         FedDST             &22.4   &55.4           &75.1           &79.9 \\ 
         FedDST$_{\mu=1.0}$ &17.3   &43.7           &71.1           &78.1     \\ 
         \hline
         FedSGC             &19.2   &\textbf{59.7}           &\textbf{77.7}  &\textbf{84.4} \\ 
         FedSGC$_{\mu=1.0}$ &19.6   &46.9           &74.4           &83.3  \\ 
        \bottomrule
    \end{tabular}
		}
		\caption{On pathological non-iid MNIST. We fix $ S=0.8, \alpha=0.5, \Delta R=20,\Delta T=20, \lambda=0.01 $}
		\label{tab:mnist_noniid}
	\end{minipage}\hfill
	\begin{minipage}{.45\linewidth}
		\centering
		\resizebox{1\columnwidth}{!}{
			    \begin{tabular}{lcccc} 
        \toprule
         \textbf{CIFAR10} &  \multicolumn{4}{c}{ Best accuracy encountered at}\\ 
         &  \multicolumn{4}{c}{cumulative upload capacity [MiB]}\\
         Method             &50     &100    &200    &400 \\
          \midrule
         FedAvg             &18.3   &22.5   &31.7   &35.2 \\ 
         FedProx            &14.9   &22.3   &24.0   &34.3 \\ 
         PruneFL            &19.3   &21.3   &30.9   &36.0     \\ 
         GraSP              &21.3   &24.3   &30.5   &34.9     \\ 
         FedDST             &28.5   &36.2   &38.6   &40.1 \\ 
         FedDST$_{\mu=0.1}$ &30.5   &35.2   &37.2   & 41.2    \\ 
         \hline
         FedSGC             &\textbf{33.3}   &\textbf{36.6}   &\textbf{41.1}    &\textbf{44.6} \\ 
         FedSGC$_{\mu=0.1}$ &32.3       &36.0       &40.2   &43.5   \\ 
        \bottomrule
    \end{tabular}
		}
		\caption{On pathological non-iid CIFAR-10. We fix $ S=0.8, \alpha=0.01, \Delta R=20,\Delta T=10, \lambda=0.01 $}
	 \label{tab:cifar10_noniid}
	\end{minipage}
\vspace{-4mm}
\end{table*}

\begin{figure}[!t]
\begin{minipage}{1\textwidth}
	\centering
		\begin{minipage}{.46\linewidth}
		\centering
		\includegraphics[width=0.89\columnwidth]{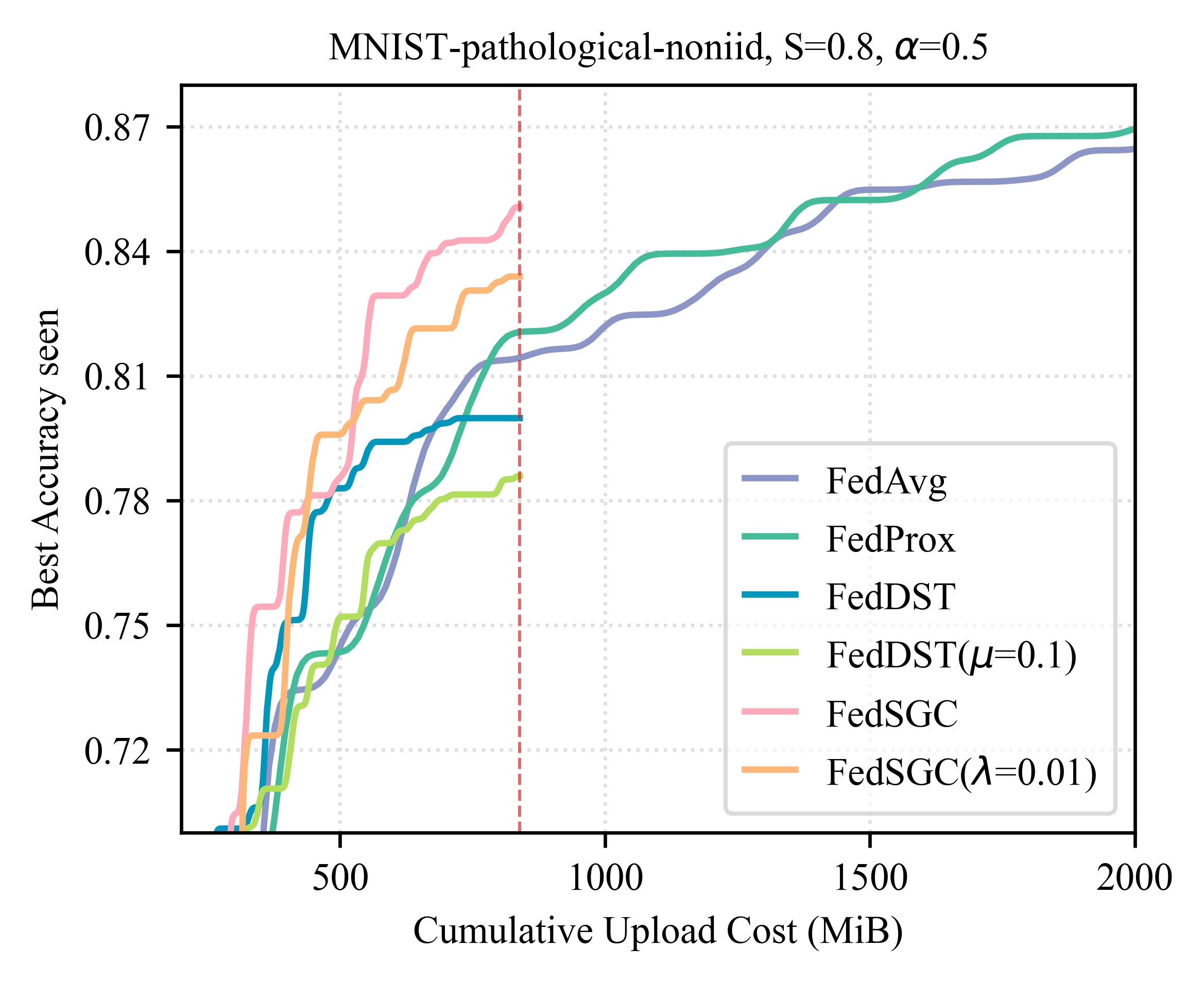}
		\vspace{-0.7mm}
		\captionof{figure}{Results on pathological non-iid MNIST dataset.}
		\label{fig:mnist_noniid_comp}
	\end{minipage}\hspace{8mm}
	\begin{minipage}{.46\linewidth}
		\centering
		\includegraphics[width=0.89\columnwidth]{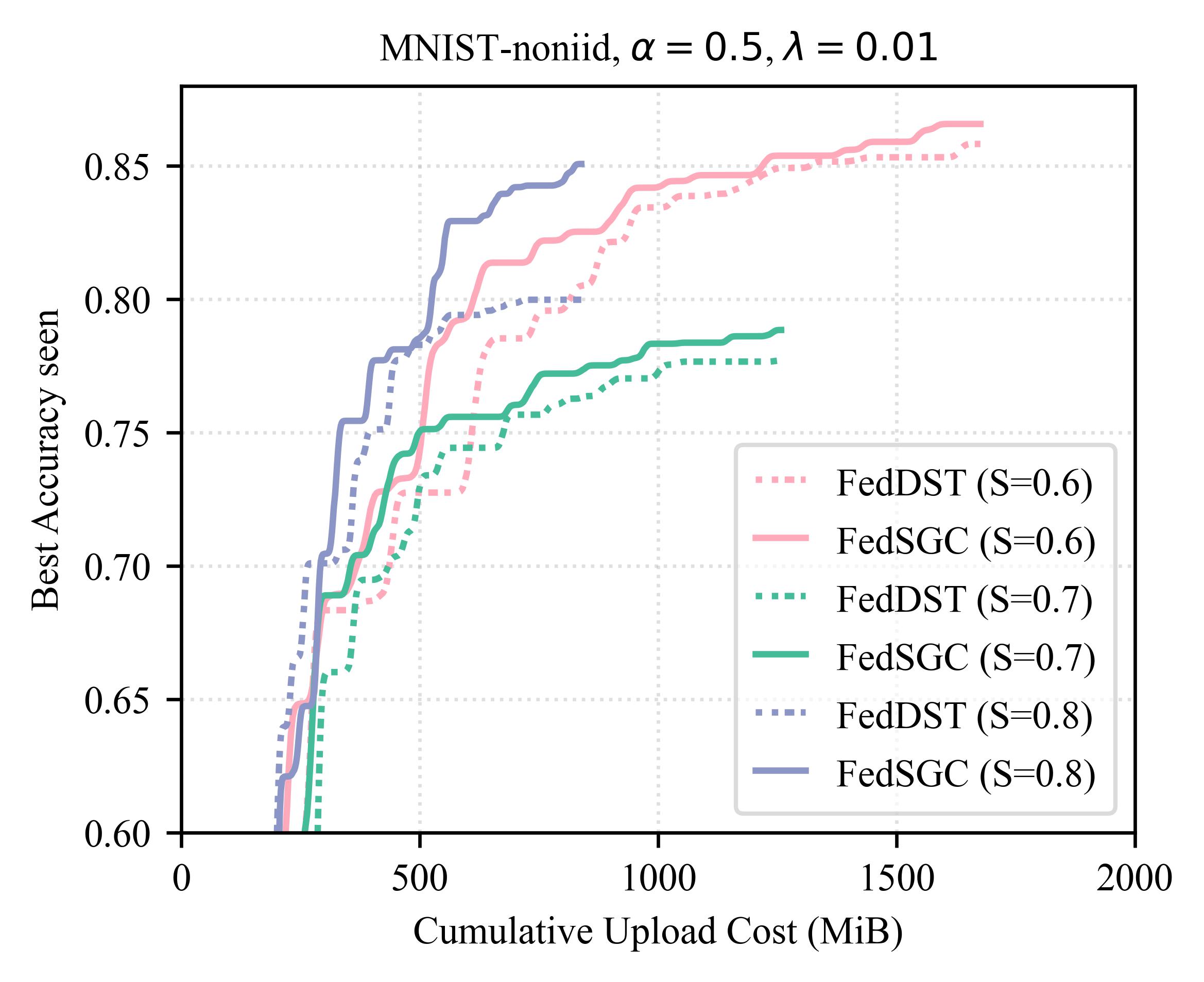}
		\vspace{-0.7mm}
		\captionof{figure}{Results on pathological non-iid MNIST dataset of different sparsity.}
		\label{fig:mnist_noniid}
	\end{minipage}
	\vspace{-4mm}
\end{minipage}
\end{figure}


\textbf{Result analysis. } {We compared the accuracy of each method on challenging pathological partitions with a fixed upload bandwidth. The results are shown in Table~\ref{tab:mnist_noniid} and Figure~\ref{fig:mnist_noniid}. PruneFL and GraSP were excluded due to their high bandwidth consumption. Our FedSGC outperformed other methods in accuracy and convergence speed, despite not excelling at the initial stage, likely due to the model’s random initialization. FedSGC quickly reached over 80\% accuracy and achieved the highest accuracy within a cumulative upload bandwidth of around 840MiB. It also proved compatible with other FL regularization terms like FedProx, achieving better results than other baselines.}


\textbf{Comparison at different Sparsity Levels. } {We tested FedSGC’s robustness and efficacy at sparsity levels 0.6, 0.7 and 0.8, maintaining other parameters and settings. Figure~\ref{fig:mnist_noniid_comp} displays the results, showing FedSGC consistently outperforming FedDST across sparsity levels, yielding superior performance under identical upload capacity. FedSGC exhibits optimal performance at higher sparsity levels, suggesting the importance of effective neuron pruning and growth, to which gradient congruity-guided FedSGC significantly contributes.}

\subsection{Results on CIFAR10}
\textbf{Implementation. } {We use the same data partition setttings as for MNIST. We follow FedAvg to use the network architecture from the TensorFlow official tutorial \cite{tfcifar} , which is a CNN with three 5x5 convolution layers (the first with 32 channels, the second and third with 64), a fully connected layer with 1024 units and ReLU activation, and a final softmax output layer. The local training is performed 20 epochs, via Adam with a learning rate of $lr=0.0001$ and batch size 50. We also incorporate the FedProx term with $\mu=0.1$ to show that our method is compatible and effective with other FL frameworks.}

\begin{figure}[!t]
\begin{minipage}{1\textwidth}
	\centering
		\begin{minipage}{.46\linewidth}
		\centering
		\includegraphics[width=0.9\columnwidth]{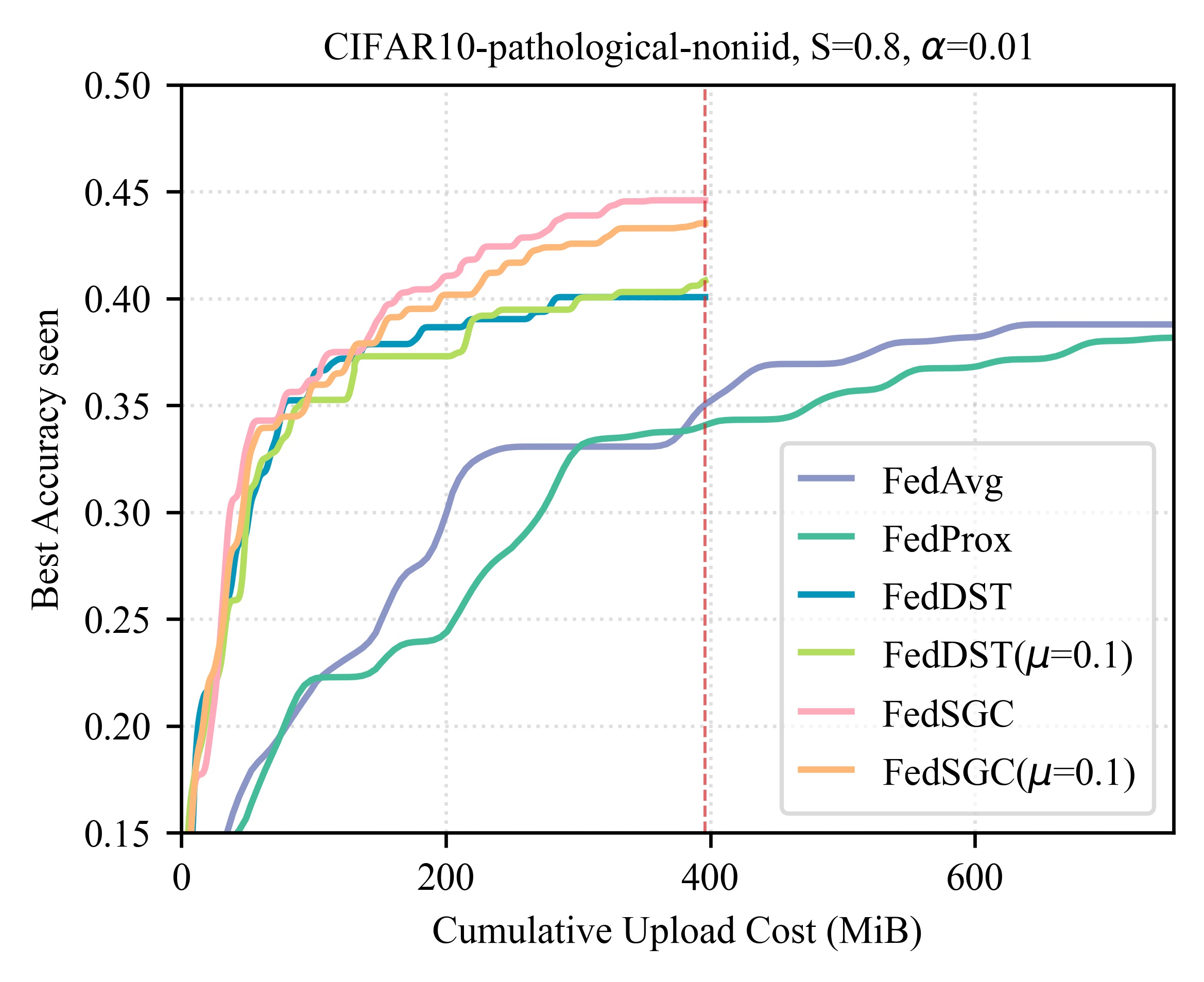}
		\vspace{-0.7mm}
		\captionof{figure}{Results on pathological non-iid CIFAR10 dataset.}
		\label{fig:cifar10-noniid}
	\end{minipage}\hspace{8mm}
	\begin{minipage}{.46\linewidth}
		\centering
		\includegraphics[width=0.9\columnwidth]{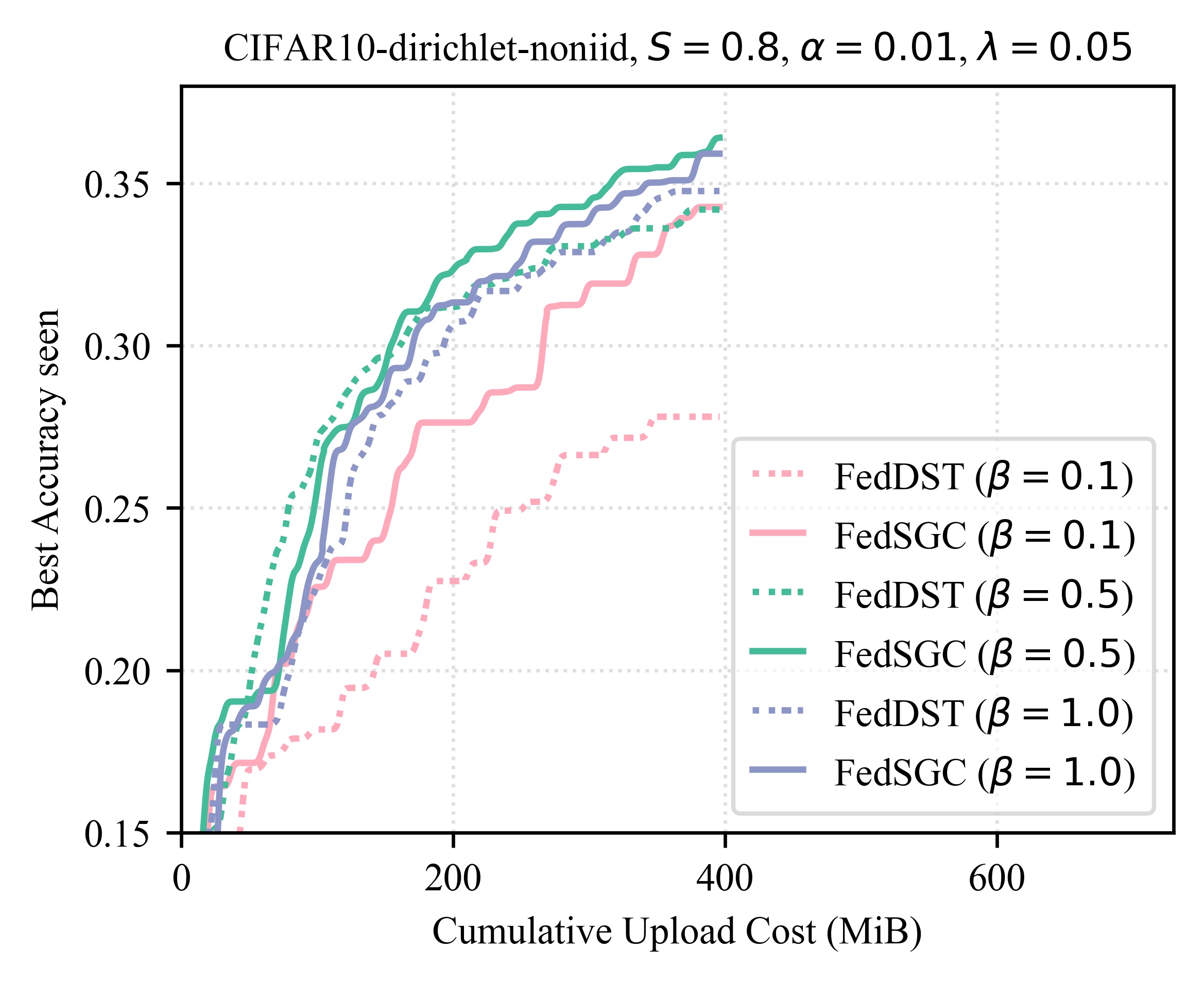}
		\vspace{-0.7mm}
		\captionof{figure}{Results on Dirichlet distributed CIFAR10 dataset of different $\beta$ value.}
		\label{fig:cifar10-dire}
	\end{minipage}
	\vspace{-4mm}
\end{minipage}
\end{figure}

\textbf{Result analysis. }{Table~\ref{tab:cifar10_noniid} and Figure~\ref{fig:cifar10-noniid} present the results. Both FedDST and FedSGC, benefiting from dynamic sparse training, show marked performance enhancements over full parameter training methods like FedAvg and FedProx. This is attributed to the sub-network ensembling effect of DST, where each client’s unique mask (network topology) represents their local data features and contributes to the global ensemble. FedSGC surpasses FedDST early on (around 40 MiB upload capacity), maintaining this lead and achieving the highest accuracy at the training’s end. This trend is consistent across all methods with FedProx terms, with our FedSGC with FedProx term ranking a close second. We roughly tuned the FedProx term weight from [0.01, 0.1, 1.0], selecting the best one, demonstrating our method’s compatibility and effectiveness with other FL algorithms.}

\textbf{Comparison at different Dirichlet parameters. } {To ascertain the resilience of our proposed FedSGC in diverse heterogeneous data environments, we conducted an evaluation using a Dirichlet-distributed CIFAR-10 dataset with varying parameters of $\beta$ (0.1, 0.5, and 1.0). A higher $\beta$ value corresponds to a larger number of locally observed classes, indicating more homogeneous client distributions. For instance, with $\beta=0.1$, the unique labels per client range from 1 to 10. As depicted in Figure \ref{fig:cifar10-dire}, FedSGC consistently outperforms FedDST, with the performance gap widening as $\beta$ decreases, indicating increased heterogeneity.}

\section{Conclusions}

In this paper, we introduce FedSGC, a novel federated sparse training scheme that seamlessly combines sparse neural networks and FL paradigms with the inspection of gradient congruity, which can effectively reduce the communication and computation costs of federated learning, while achieving competitive performance under heterogeneous data distributions. We have evaluated FedSGC on various datasets and compared it with several state-of-the-art baselines, showing that FedSGC can consistently achieve comparable or better results with significant communication savings. Moreover, we have demonstrated that FedSGC is compatible with other popular federated optimization frameworks such as FedProx, and can maintain its effectiveness with minimal overhead.
\section*{Acknowledgments}
This work was supported in part by the Hong Kong Innovation and Technology Commission (InnoHK Project CIMDA), in part by Research Grant Council General Research Fund 11203220, and in part by CityU Strategic Interdisciplinary Research Grant 7020055.

\bibliography{iclr2024_conference}
\bibliographystyle{iclr2024_conference}

\appendix
\section{Appendix}

\subsection{Communication and Computation Savings}
\textbf{Communication Analysis. } The sparse training nature of FedSGC allows it to conserve substantial communication bandwidth, both in terms of upload and download, when compared to the full training methods like FedAvg. Let’s denote the total number of network parameters as n, each taking up 4 bytes (32 bits). Given a target sparsity S, the average upload and download cost for FedSGC in most communication rounds is $32(1-S)n$. With a mask re-adjust round frequency of $\Delta R$, FedSGC incurs an additional cost of $2(1-S)n$ during the re-adjusting rounds for distributing the global direction map and the newly adjusted mask. Therefore, the average download communication cost for FedSGC is $(32(1-S)+\frac{2}{\Delta R})n$.

\textbf{Computation Analysis. } FedSGC also significantly reduces local computational workloads by maintaining sparse networks throughout the FL process. In most rounds, FedSGC does not require full dense training. In terms of FLOP savings, this allows us to bypass most of the FLOPs in both training and inference, proportional to the model’s sparsity. Only few epochs of full training are needed in the re-adjust round for neuron growth. This cost can be further optimized by limiting the parameters exploration space to a sparsity lower than the target sparsity only (e.g., $\frac{S}{2}$), instead of exploring in the full parameters space (i.e., zero sparsity).

\subsection{Results on PACS}
Furthermore, to evaluate the robustness of our method against data heterogeneity, we also use the PACS dataset, which is a mainstream benchmark for domain generalization tasks. This dataset contains four different domains (i.e., Photo, Art Painting, Cartoon, Sketch) with the same seven label classes (i.e., Dog, Elephant, Giraffe, Guitar, Horse, House, Person). The data distribution among different domains is significantly different, so we can test the global model’s generalization ability against the data heterogeneity by using the leave-one-domain-out strategy, where we use three domains as three clients for federated training and the remaining one domain for testing.
\begin{figure}
	[t]
	\centering 
	\includegraphics[width=1\columnwidth]{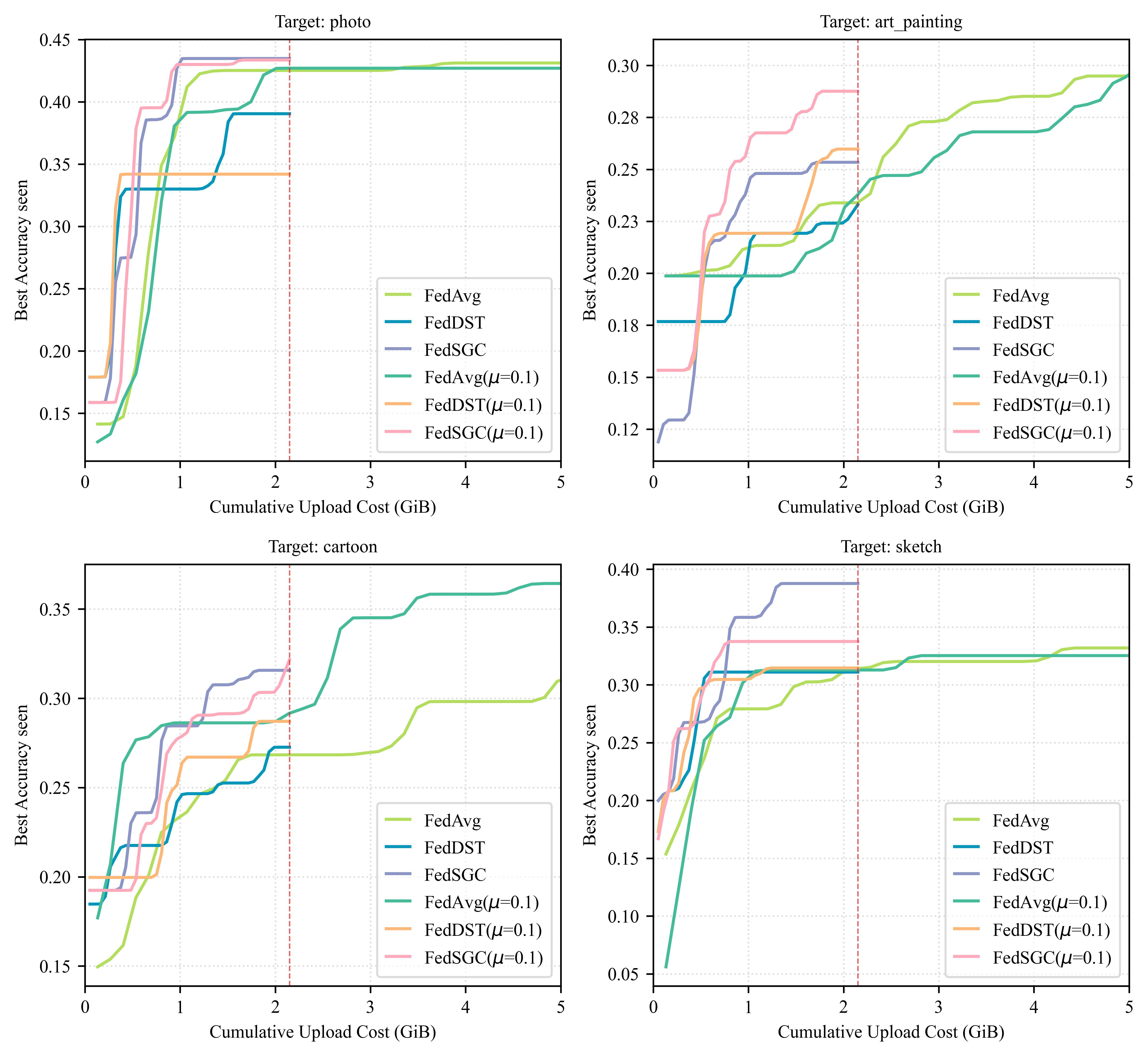} %
	\vspace{-7mm}
        \caption{Results on PACS dataset.}
	\label{fig:pacs-sperate}
\end{figure}
\begin{table}
[t]
\caption{Accuracy of FedSGC and other baseline methods given cumulative upload bandwidth limits, on PACS dataset. We fix $S$ = 0.6 for all sparse methods, $\alpha=0.1,\Delta R=7$ for FedDST and FedSGC methods. $\lambda=0.005$ for FedSGC. We set $\mu=0.1$ for all experiments with FedProx terms. The same settings are applied for all three experiments of different target domains.}
    \centering
    \begin{tabular}{lcccc} 
        \toprule
         \textbf{PACS} &  \multicolumn{4}{c}{Best accuracy encountered at}\\ 
         &  \multicolumn{4}{c}{cumulative upload capacity [GiB]}\\
         Method             &0.5     &1.0    &1.5    &2.0 \\
          \midrule
          \midrule
          \multicolumn{5}{c}{\textbf{Photo}}  \\ \hline
FedAvg              &17.5 &42.2 &42.5 &42.5 \\
FedProx             &17.5 &39.2 &39.4 &42.7 \\
PruneFL             &16.4 &26.1 &30.1 &33.3 \\
GraSP               &20.3 &32.4 &34.5 &36.7\\
FedDST              &33.0 &33.0 &39.0 &39.0 \\
FedDST$_{\mu=0.1}$  &34.2 &34.2 &34.2 &34.2 \\\hline
FedSGC              &27.5 &\textbf{43.5} &\textbf{43.5} &\textbf{43.5} \\
FedSGC$_{\mu=0.1}$  &\textbf{39.5} &43.0 &43.0 &43.4 \\
        \bottomrule
         \multicolumn{5}{c}{\textbf{Art Painting}}  \\ \hline
FedAvg              &20.2 &21.3 &21.3 &23.4 \\
FedProx             &19.9 &19.9 &19.9 &23.6 \\
PruneFL             &14.3 &17.1 &17.3 &21.3 \\
GraSP               &15.5 &16.4 &18.4 &24.5 \\
FedDST              &17.7 &21.9 &21.9 &22.4 \\
FedDST$_{\mu=0.1}$  &21.4 &21.9 &21.9 &26.0 \\\hline
FedSGC              &20.2 &24.8 &24.8 &25.3 \\
FedSGC$_{\mu=0.1}$  &22.8 &\textbf{26.8} &\textbf{27.8} &\textbf{28.8} \\
         \bottomrule
         \multicolumn{5}{c}{\textbf{Cartoon}}  \\ \hline
FedAvg              &19.5 &23.5 &25.2 &26.8 \\
FedProx             &27.7 &\textbf{28.6} &28.6 &28.6 \\
PruneFL             &15.6 &21.5 &24.3 &26.4 \\
GraSP               &17.3 &24.9 &25.1 &27.5 \\
FedDST              &21.8 &24.7 &25.3 &27.3 \\
FedDST$_{\mu=0.1}$  &20.0 &26.7 &26.7 &28.7 \\\hline
FedSGC              &\textbf{23.6} &28.5 &\textbf{30.8} &\textbf{31.6} \\
FedSGC$_{\mu=0.1}$  &19.2 &27.9 &29.1 &30.3 \\
        \bottomrule
         \multicolumn{5}{c}{\textbf{Sketch}}  \\ \hline
FedAvg              &23.2 &27.9 &30.2 &31.4 \\
FedProx             &26.4 &31.3 &31.3 &31.3 \\
PruneFL             &22.1 &25.3 &29.3 &30.5 \\
GraSP               &24.3 &31.2 &32.6 &33.9 \\
FedDST              &31.1 &31.1 &31.1 &31.1 \\
FedDST$_{\mu=0.1}$  &29.7 &30.5 &31.5 &31.5 \\\hline
FedSGC              &26.8 &\textbf{35.8} &\textbf{38.8} &\textbf{38.8} \\
FedSGC$_{\mu=0.1}$  &\textbf{30.1} &33.7 &33.7 &33.7 \\
        \bottomrule
    \end{tabular}
    \label{tab:pacs-avg}
\end{table}
\textbf{Implementation. } {We follow the settings of previous works which also adopt PACS in federated learning experiments \cite{nguyen2022fedsr, zhang2023federated}. We use ResNet-18 as the backbone, SGD with learning rate 0.001 as the optimizer for clients' training. The target sparsity $S$ to 0.6. We iteratively pick one domain out as the target domain for global model evaluation, and make the remaining 3 domains as clients to join the federated learning process for 40 rounds, 2 epochs each round. Again, we compare all methods with the FedProx term using weight $\mu=0.1$ for all experiments, after rough tuning $\mu$ from [0.01, 0.1, 1.0].}

\textbf{Result Analysis. } {We conduct four separate experiments, each with a different domain as the target domain for testing. The full results of the four experiments are reported in Table ~\ref{tab:pacs-avg}, and the four line graphs showing the best accuracy achieved along the training process are presented in Figure ~\ref{fig:pacs-sperate}. As we can see, in all experiments, FedSGC and FedSGC with FedProx term can consistently outperform the other baselines, ranking first or second. Especially in the experiment with the target domain Sketch, which is known to have the largest domain gap with the other three domains, FedSGC outperforms the other dynamic sparse training baseline, FedDST, significantly by 7.7\%. We also find that the sparse training with target domains of ArtPainting and Cartoon benefits most from the FedProx term, as both FedDST and FedSGC with the prox term outperform the ones without. Surprisingly, we find that in the target-Photo and target-Sketch experiments, FedSGC can achieve even better performance using less upload capacity compared with full parameter training methods (FedAvg and FedProx). In all, our FedSGC shows greate effectiveness in realistic and challenging datasets PACS, beyond the difficulty of MNIST and CIFAR-10.}

\subsection{Details of Algorithm}
\begin{algorithm}[t]
	\centering
	\caption{Federated Dynamic Sparse Training}
	\label{alg:fds}
	\begin{small}
		\begin{algorithmic}[1] %
			\REQUIRE ~~~~Clients $[N]$ with local datasets $D_i$ \\
			\quad~~~~~~$S = \{s_1, \ldots, s_l\}$ $\to$ Sparsities by layer \\
			\quad~~~~~~$\Delta R$, $\Delta T$, $T_{end}$, $\alpha_r$ $\to$ Update schedule \\
			\quad~~~~~~$\mathcal{D}=\{D_s\}_{s=1}^{S}$ $\to$ training dataset with $C$ classes.\\

			\ENSURE ~$\mathcal{\theta^R}, m^R$ $\to$ The final produced global model parameters and mask\\
\STATE $\triangleright$ \textit{\CommetBlue{Main procedure starts}}
\STATE Initialize server model $(\theta_1, m_1, d_0)$ at sparsity $\|m_1\|_0 = S, d_1=0$;\\
\FOR {each round $r \in [R]$}
\STATE Sample clients $C_r \subset [N]$; \\
\STATE Transmit the server model $(\theta_r, m_r, d_r)$ to all clients $c \in C_r$;
\FOR {each client $c \in C_r$ do in parallel}
\STATE Receive $(\theta_r^c, m_r^c,d_r) \gets (\theta_r, m_r, d_r)$ from the server;
\FOR {each epoch $e \in [E]$}
\STATE $e_c \gets e_c + 1$
\STATE Sample a mini-batch $B$ from $D_c$, start the training iteration $t$ on $B$;
\STATE Perform one step of local training of local sparse network $\theta_r^c \odot m_r^c$;
\IF {$r \bmod \Delta R = 0$ and $r < R_{end}$ and $e_c \bmod \Delta T=0$ and $e_c < T^{end}_c$ }
\STATE Perform layer-wise magnitude pruning $(\theta_r^c, m_r^c) \gets \text{prune}(\theta_r^c; S, \sigma_c, d_r)$ to attain sparsity distribution $\overline{S_c}$ using Eq.~\ref{eq:drop_guided} and Eq.~\ref{eq:drop_rest};
\STATE Perform layer-wise loss gradient magnitude growth $(\theta_r^c, m_r^c) \gets \text{grow}(\theta_r^c; S, \sigma_c, d_r)$ to restore sparsity distribution $S$ using Eq.~\ref{eq:grow_guided} and Eq.~\ref{eq:grow_rest};
\ENDIF
\ENDFOR
\STATE Transmit the new $(\theta_r^c, m_r^c)$ to the server;
\ENDFOR
\STATE Receive the updated client-local networks and masks $(\theta_r^c, m_r^c)$ from clients $c \in C_r$;
\STATE Aggregate and prune global networks to get  $(\theta_{r+1}, m_{r+1})$ using Eq.~\ref{eq:central_agg} to attain sparsity distribution $S$;
\STATE Get global pseudo-gradient direction map \\ $d_{r+1}= sign(\theta_{r+1} - \theta_r)$ for next round training.
\ENDFOR
		\end{algorithmic}
	\end{small}
\end{algorithm}
\end{document}

%% file: math_commands.tex
\usepackage{amsmath,amsfonts,bm}

\def\eqref#1{equation~\ref{#1}}

\def\1{\bm{1}}

\DeclareMathAlphabet{\mathsfit}{\encodingdefault}{\sfdefault}{m}{sl}
\SetMathAlphabet{\mathsfit}{bold}{\encodingdefault}{\sfdefault}{bx}{n}

